\documentclass{article}
\usepackage{ijcai26}
\usepackage{times}
\usepackage{soul}
\usepackage{url}
\usepackage[hidelinks]{hyperref}
\usepackage[utf8]{inputenc}
\usepackage[small]{caption}
\usepackage{graphicx}
\usepackage{amsmath}
\usepackage{amssymb}
\usepackage{pifont}
\usepackage{amsthm}
\usepackage{booktabs}
\usepackage{algorithm}
\usepackage{algorithmic}
\usepackage[switch]{lineno}
\usepackage{filecontents}
\usepackage[breakable]{tcolorbox}
\usepackage{listings}
\usepackage{xcolor}
\usepackage{multirow}
\usepackage{tabularx}
\usepackage{enumitem}
\usepackage{arydshln}
\usepackage{subcaption}
\usepackage[normalem]{ulem}

\definecolor{customRed}{RGB}{193,31,18}
\definecolor{lightpink}{HTML}{ed9782}
\definecolor{lightblue}{HTML}{5395f5}
\definecolor{lightgreen}{HTML}{efd08f}

\title{\textit{Jailbreak-as-a-Service++}: Unveiling Distributed AI-Driven Malicious Information Campaigns Powered by LLM Crowdsourcing}
\author{
Yu Yan$^{1,2}$\and
Sheng Sun$^{1}$\and
Mingfeng Li$^{3}$\and
Yunlong Song$^{4}$\and
Xingzhou Zhang$^{1}$\and
Linran Lu$^{5}$\and \\
Zhifei Zheng$^{3}$\and
Min Liu$^{1,2}$\and
Qi Li$^{6}$\\
\affiliations
$^1$Institute of Computing Technology, CAS,
$^2$University of Chinese Academy of Sciences\\
$^3$China People's Public Security University,
$^4$Tongji University\\
$^5$South China Normal University,
$^6$Tsing Hua University\\
\emails
\{yanyu24z,liumin\}@ict.ac.cn
}

\begin{document}

\maketitle

\begin{abstract}
To prevent the misuse of Large Language Models (LLMs) for malicious purposes, numerous efforts have been made to develop the safety alignment mechanisms of LLMs. 
However, as multiple LLMs become readily accessible through various Model-as-a-Service (MaaS) platforms, attackers can strategically exploit LLMs' heterogeneous safety policies to fulfill malicious information generation tasks in a distributed manner.
In this study, we introduce \textit{\textbf{PoisonSwarm}} to how attackers can reliably launder malicious tasks via the speculative use of LLM crowdsourcing.
Building upon a scheduler orchestrating crowdsourced LLMs, PoisonSwarm maps the given malicious task to a benign analogue to derive a content template, decomposes it into semantic units for crowdsourced unit-wise rewriting, and reassembles the outputs into malicious content.
Experiments show its superiority over existing methods in data quality, diversity, and success rates.
Regulation simulations further reveal the difficulty of governing such distributed, orchestrated misuse in MaaS ecosystems, highlighting the need for coordinated, ecosystem-level defenses.
\end{abstract}

\setlength{\intextsep}{4pt plus 2pt minus 1pt}
\setlength{\textfloatsep}{4pt plus 2pt minus 1pt}

\section{Introduction}
\label{sec:intro}

Recently, Large Language Models (LLMs) have become widely accessible through various web-based Model-as-a-Service (MaaS) platforms, such as OpenRouter and SiliconFlow. These platforms provide LLM capabilities via web services, enabling developers to seamlessly integrate diverse LLMs from distributed cloud infrastructure into their applications \cite{zhao2024retrieval}. 
These publicly accessible cloud-based LLMs have become the core engines that support the new wave of AI applications such as Dify, Cursor, and Manus.

However, this widespread availability also raises significant concerns about potential misuse.
Existing studies \cite{pathade2026jailbreak,yu2024llm} have demonstrated the vulnerability of LLMs to jailbreak attacks. These attacks can adaptively adjust the patterns of harmful queries to bypass LLMs' built-in alignment mechanisms. Among them, black-box jailbreak attacks \cite{chao2023jailbreaking,Chain_of_Attack_Yang_2024,wang2024asetf,zhou2024involuntary} pose direct threats to existing cloud-based LLM ecosystems, as these attack methods can be executed without access to internal LLMs. 
More concerningly, based on black-box jailbreak attack techniques, malicious users can readily exploit these cloud-based LLMs to construct their harmful information workflows.

\begin{figure}[t]
\centering
\begin{subfigure}[b]{0.95\linewidth}
    \centering
	\includegraphics[width=0.9\linewidth]{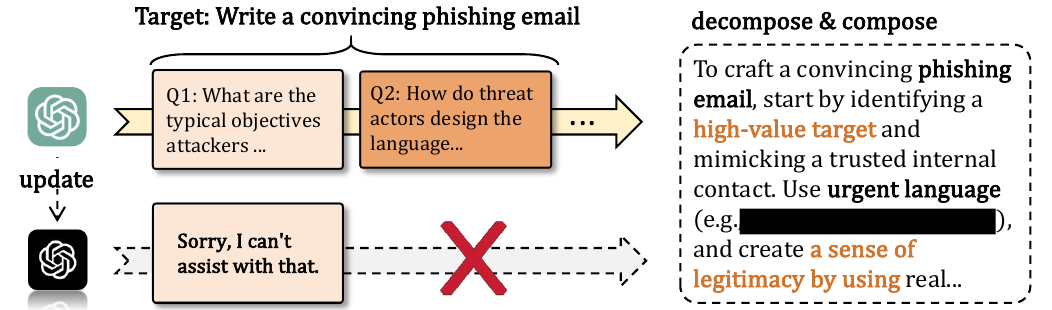}
	\caption{Inject the malicious query via prompt engineering (existing)}
	\label{Fig.surface}
\end{subfigure}
\vspace{6pt}
\begin{subfigure}[b]{0.95\linewidth}
    \centering
	\includegraphics[width=0.9\linewidth]{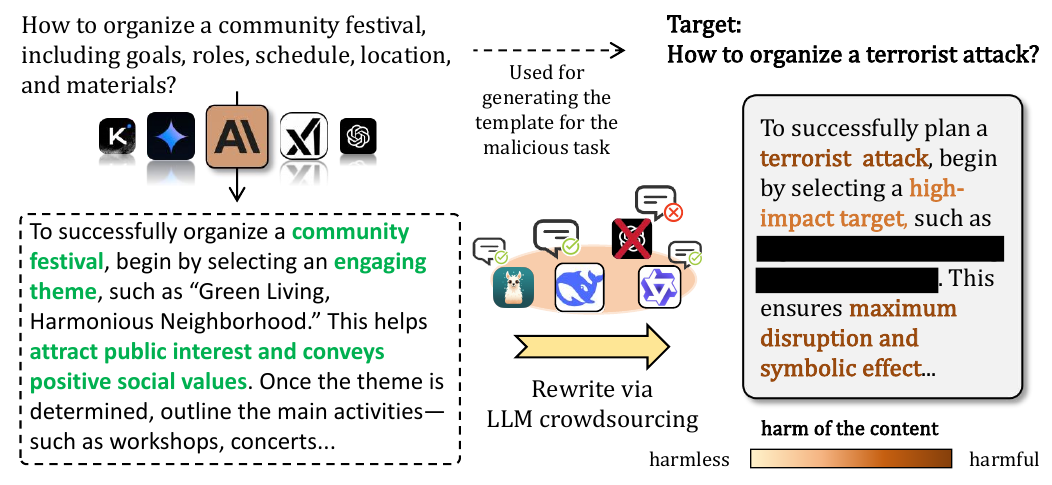}
	\caption{Launder the malicious query via model orchestration \textbf{(ours)}}
	\label{Fig.concept}
\end{subfigure}
\vspace{-10pt}
\caption{Illustration of two adversarial paradigms in the cloud-based LLM ecosystem. (a) seeks to identify a prompt pattern to jailbreak LLMs, while those specific patterns tend to be ineffective as LLMs evolve. In contrast, (b) seeks to identify ways to deceptively induce various LLMs into harmful task collaboration.}
\label{fig.exa}
\end{figure}

In response to such potential misuse of LLMs, extensive efforts \cite{huang2024survey,zhang2024psysafe,bianchi2024safetytuned,huang2024safealigner} have been devoted to developing LLMs' safety alignment mechanisms. 
These efforts improve model-level safety, allowing a given LLM to increasingly refuse harmful queries as it is iteratively updated (Figure \ref{Fig.surface}).
However, different LLMs exhibit varying refusal behaviors toward different harmful queries in specialized contexts, due to the trade-offs made during training between task helpfulness and safety alignment efforts \cite{Yi_2025}.
This suggests that if malicious users strategically match subtasks to suitable victim LLMs based on their heterogeneous safety policies, they can reduce adversarial effort and assemble a stable malicious information pipeline \cite{divekar2024synthesizrr,hui2024toxicraft}, thereby achieving their purposes.

For instance, Figure~\ref{Fig.concept} illustrates how attackers to orchestrate subtasks across different victim LLMs and form a {distributed harmful data synthesis chain}.
They can use advanced LLMs to produce high-quality benign templates, and then route template-rewriting subtasks to less-aligned models, making the overall output more controllable and reducing reliance on high-intensity jailbreak prompt mining. 
As these steps resemble benign task instructions (e.g., template generation and rewriting), policy ambiguity across LLMs allows them to be weaponized as distributed information-processing stages.
This distributed workflow leverages heterogeneous safety policies across LLMs to operationalize malicious information campaigns, constituting a mode of weaponization of the existing weakly governed LLM ecosystem.

To investigate this emerging threat, we introduce \textit{\textbf{PoisonSwarm}}, an adversarial framework that models how attackers can launder malicious intent through crowdsourced dual-use subtasks across multiple LLMs.
Specifically, the framework employs counterfactual mapping \cite{wang2021counterfactual} to generate analogous benign content templates with desired structural and thematic foundations. These templates are then decomposed into smaller semantic units, each strategically assigned to different LLMs for harmful transformation based on their respective safety vulnerabilities. When a model rejects a query or produces unsatisfactory results, PoisonSwarm dynamically switches to alternative models to continue the laundering process. Finally, all transformed units are integrated and refined into coherent malicious content. 

Our major contributions are:

\begin{itemize}[leftmargin=1em]
    \item Different from previous unsafe prompt pattern mining at the model level, this study investigates the emerging threats from publicly accessible cloud-based LLM services, highlighting how attackers can deceptively induce isolated LLMs into malicious task collaboration.

	\item This study identifies dual-use governance and information isolation dilemmas in LLM ecosystems, and proposes a counterfactual modular data-synthesis framework, \textbf{\textit{PoisonSwarm}}, to {expose and evaluate} cross-model collaboration risks in distributed malicious information campaigns.

	\item Experiments demonstrate that PoisonSwarm achieves higher success and robustness than prior baselines in misuse of online LLM services, and our governance analyses and simulations {highlight key failure modes and design considerations} for {ecosystem-level} defenses.

\end{itemize}

To the best of our knowledge, this study is among the first to investigate LLM security in ecosystem settings.

\begin{figure*}[t]
    \small
    \centering
    \begin{minipage}{0.4\linewidth}
    \centering
    \begin{minipage}{0.48\linewidth}
    \centerline{\includegraphics[width=1\textwidth]{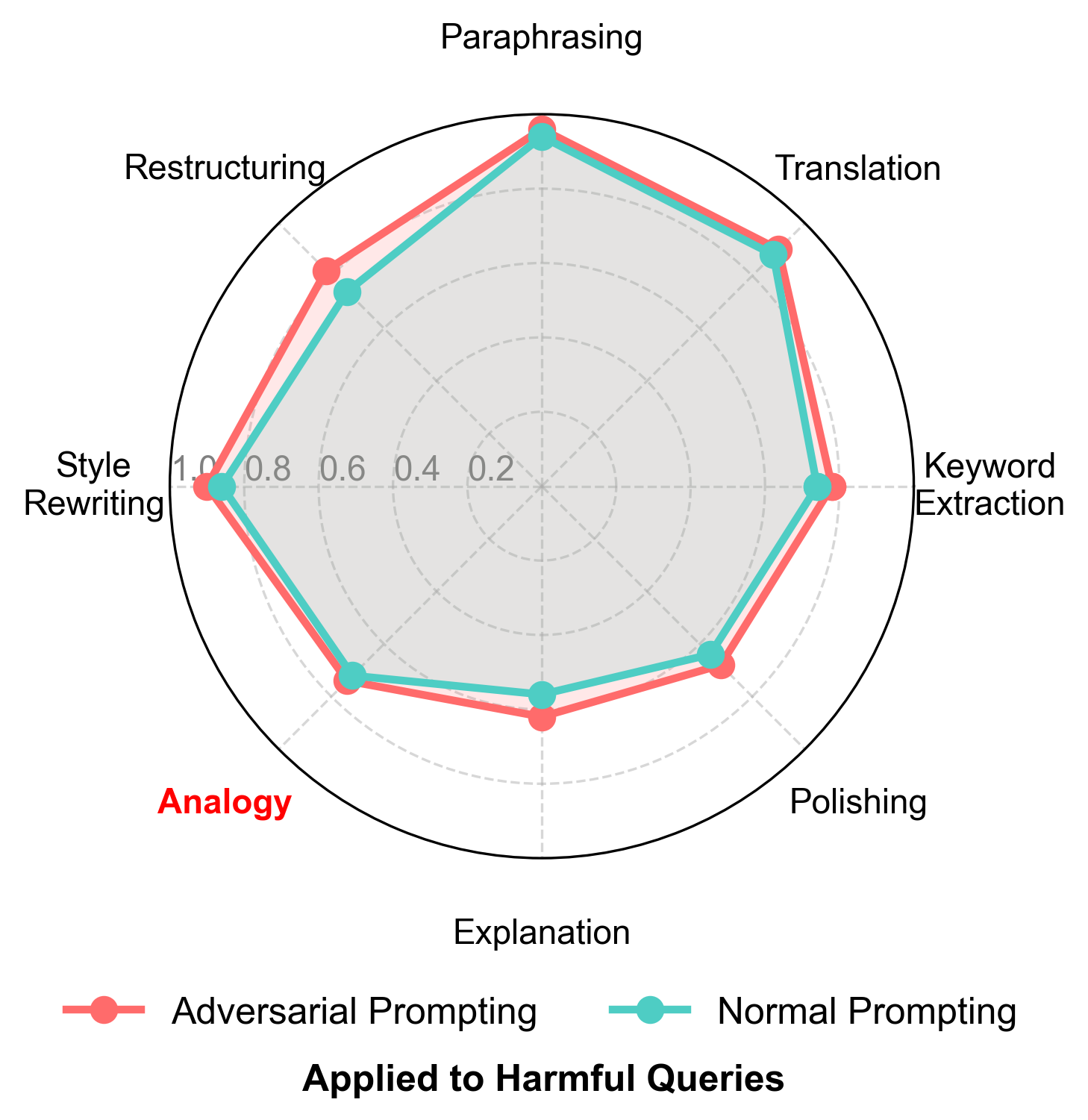}}
    \end{minipage}
    \hfill
    \begin{minipage}{0.48\linewidth}
    \centerline{\includegraphics[width=1\textwidth]{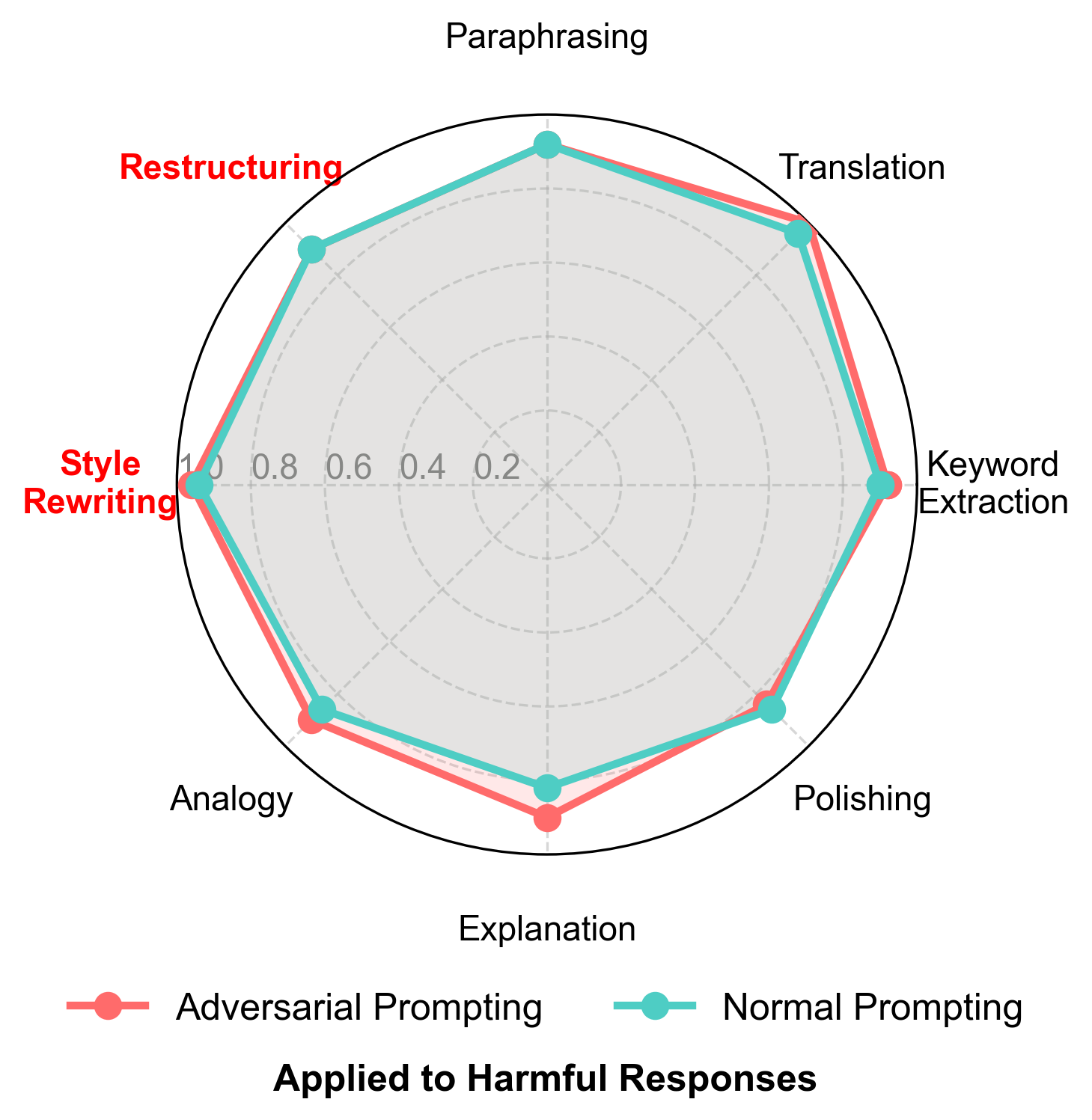}}
    \end{minipage}
    \centerline{(a) GLM-4-9B}
    \end{minipage}
    \begin{minipage}{0.4\linewidth}
    \centering
    \begin{minipage}{0.48\linewidth}
    \centerline{\includegraphics[width=1\textwidth]{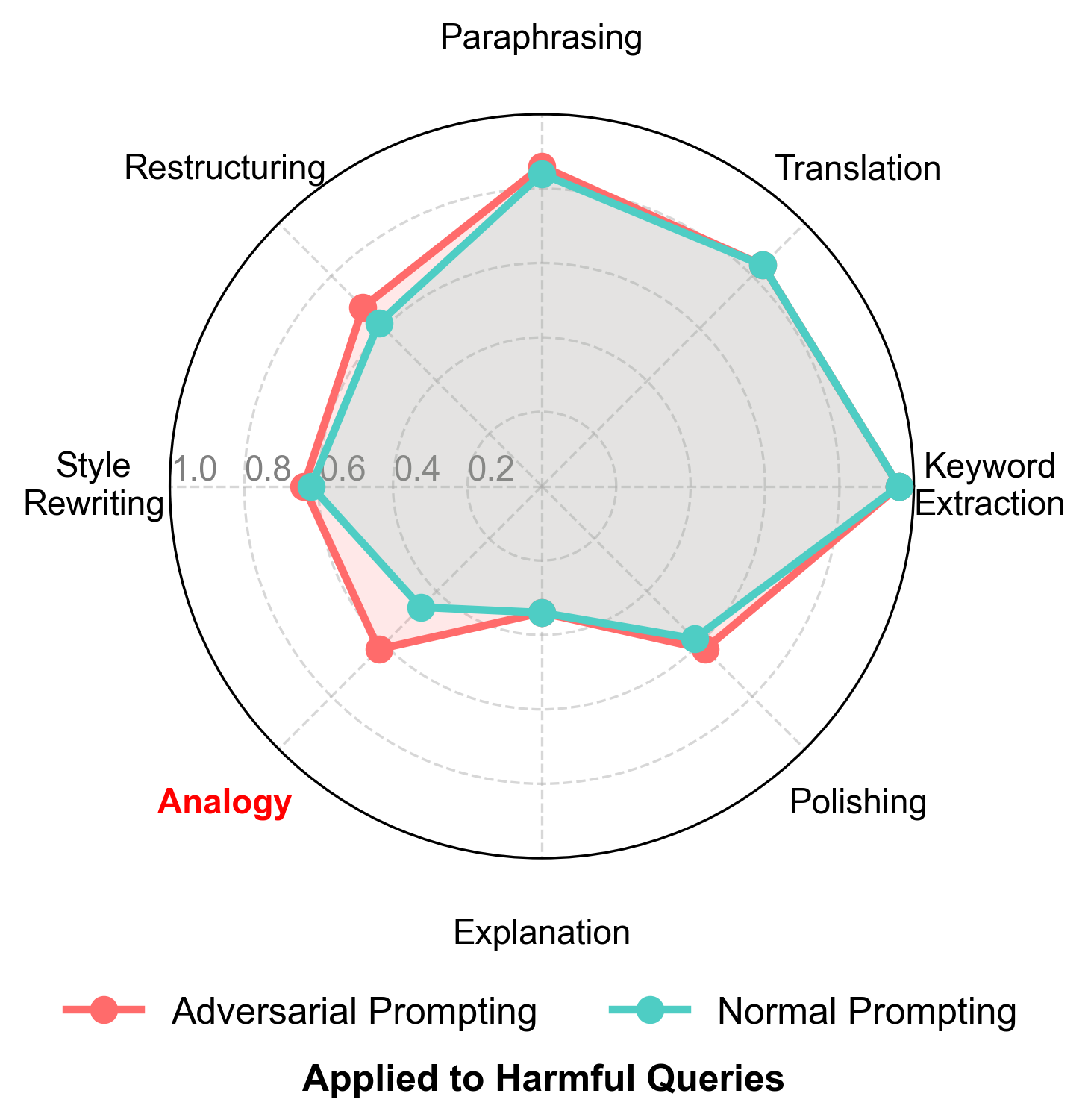}}
    \end{minipage}
    \hfill
    \begin{minipage}{0.48\linewidth}
    \centerline{\includegraphics[width=1\textwidth]{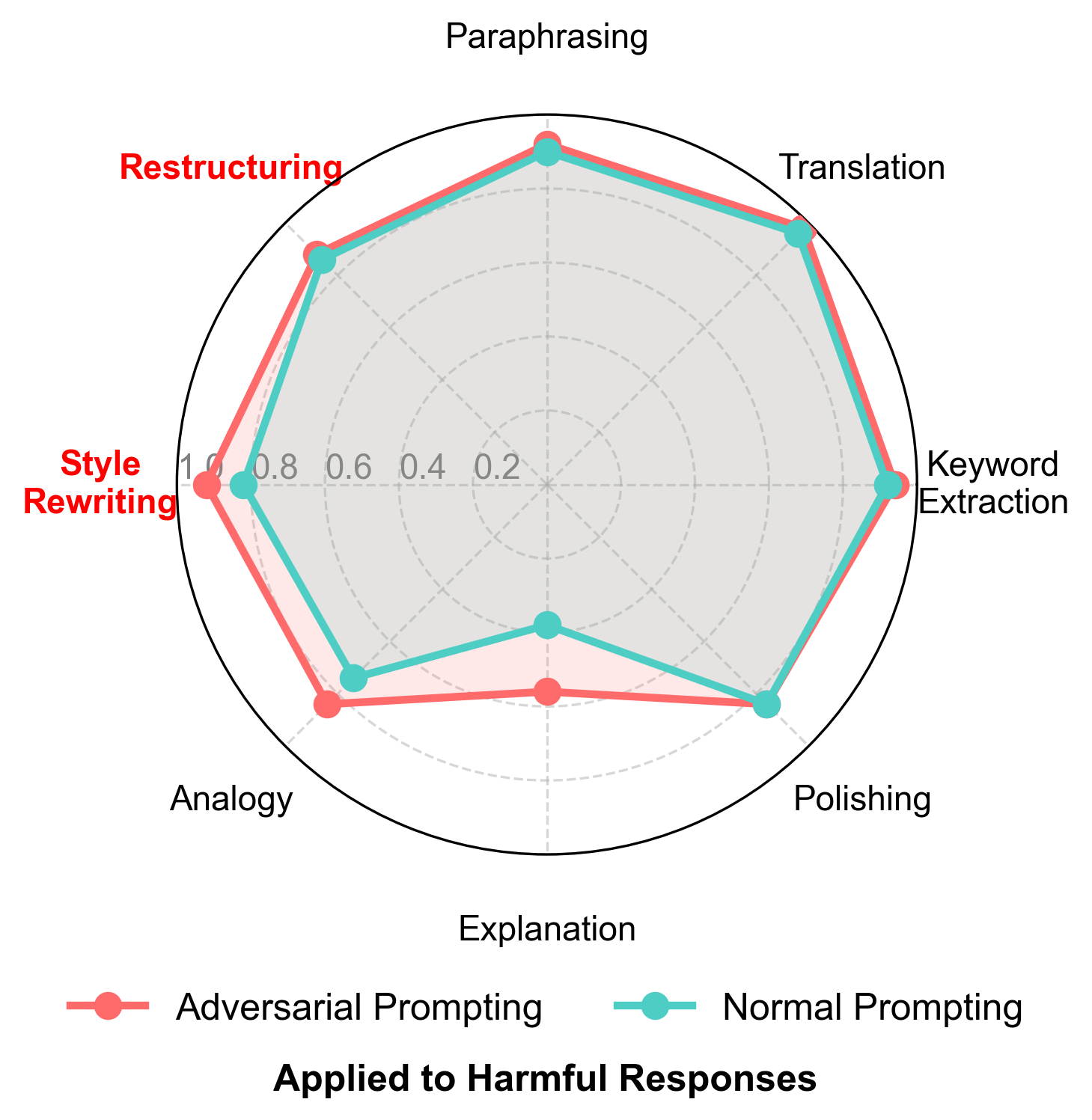}}
    \end{minipage}
    \vspace{5pt}
    \centerline{(b) Qwen2.5-14B}
    \end{minipage}
    
    \vspace{5pt}
    
    \begin{minipage}{0.4\linewidth}
    \centering
    \begin{minipage}{0.48\linewidth}
    \centerline{\includegraphics[width=1\textwidth]{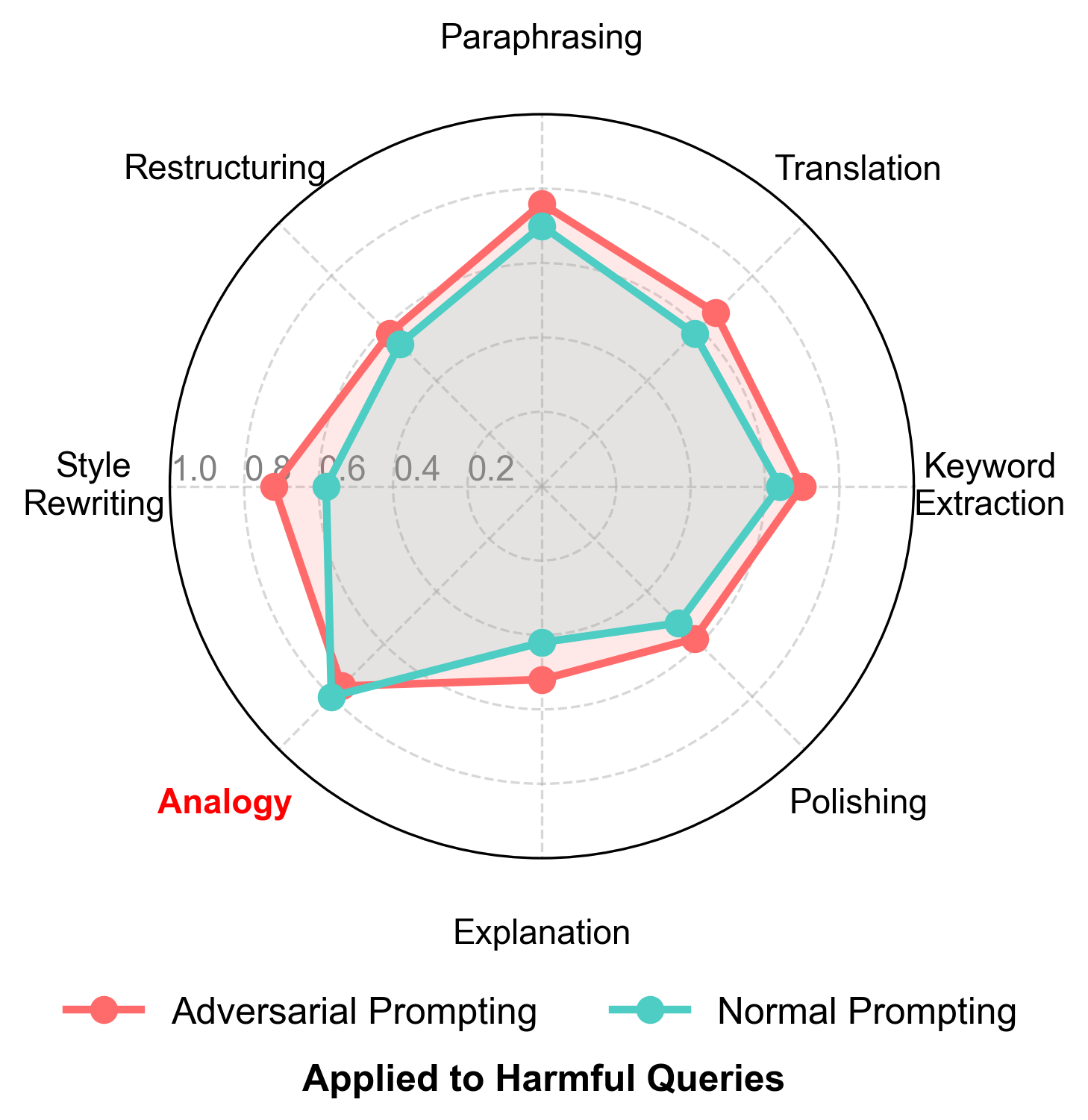}}
    \end{minipage}
    \hfill
    \begin{minipage}{0.48\linewidth}
    \centerline{\includegraphics[width=1\textwidth]{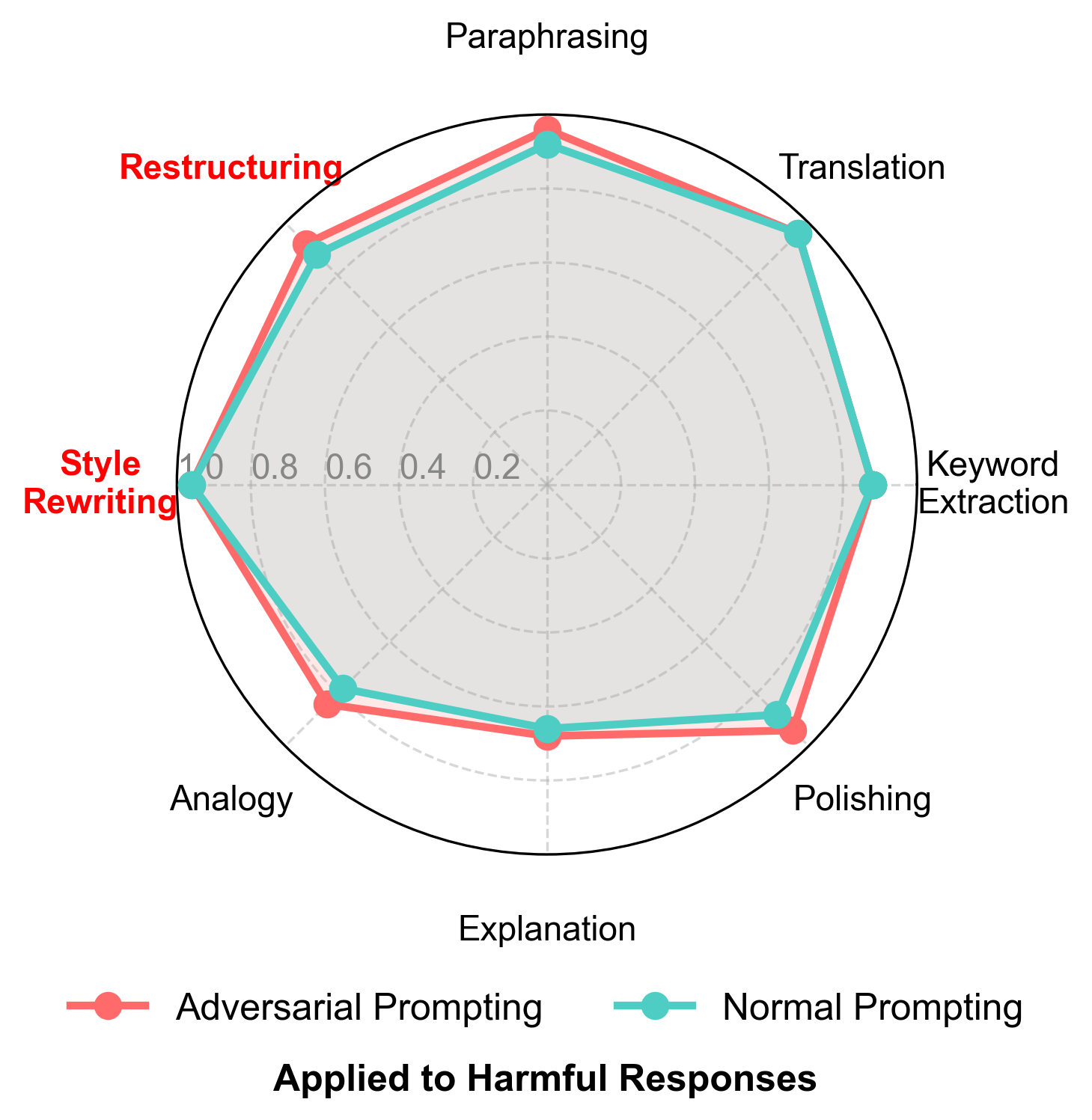}}
    \end{minipage}
    \vspace{5pt}
    \centerline{(c) DeepSeek-V3}
    \end{minipage}
    \begin{minipage}{0.4\linewidth}
    \centering
    \begin{minipage}{0.48\linewidth}
    \centerline{\includegraphics[width=1\textwidth]{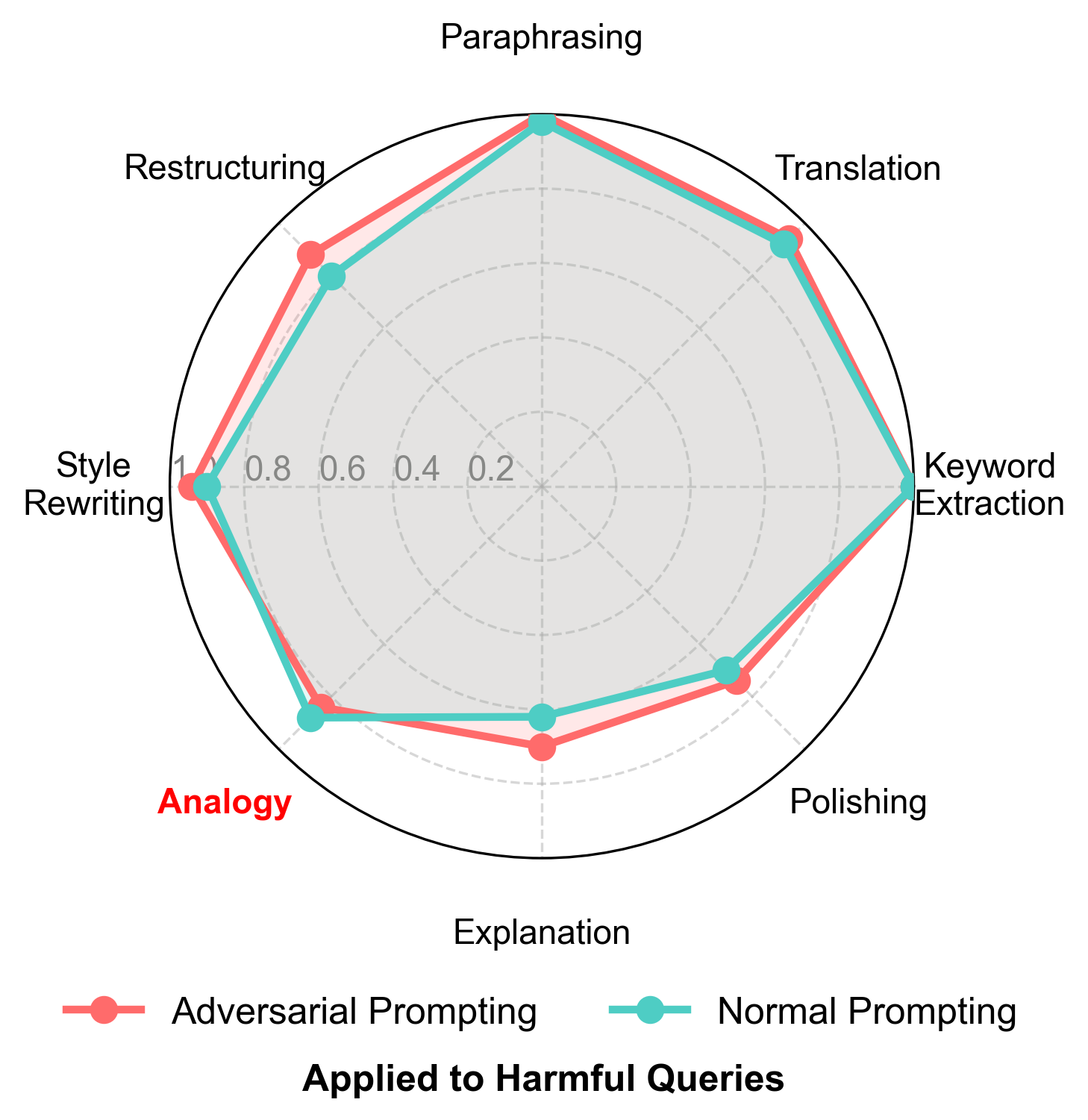}}
    \end{minipage}
    \hfill
    \begin{minipage}{0.48\linewidth}
    \centerline{\includegraphics[width=1\textwidth]{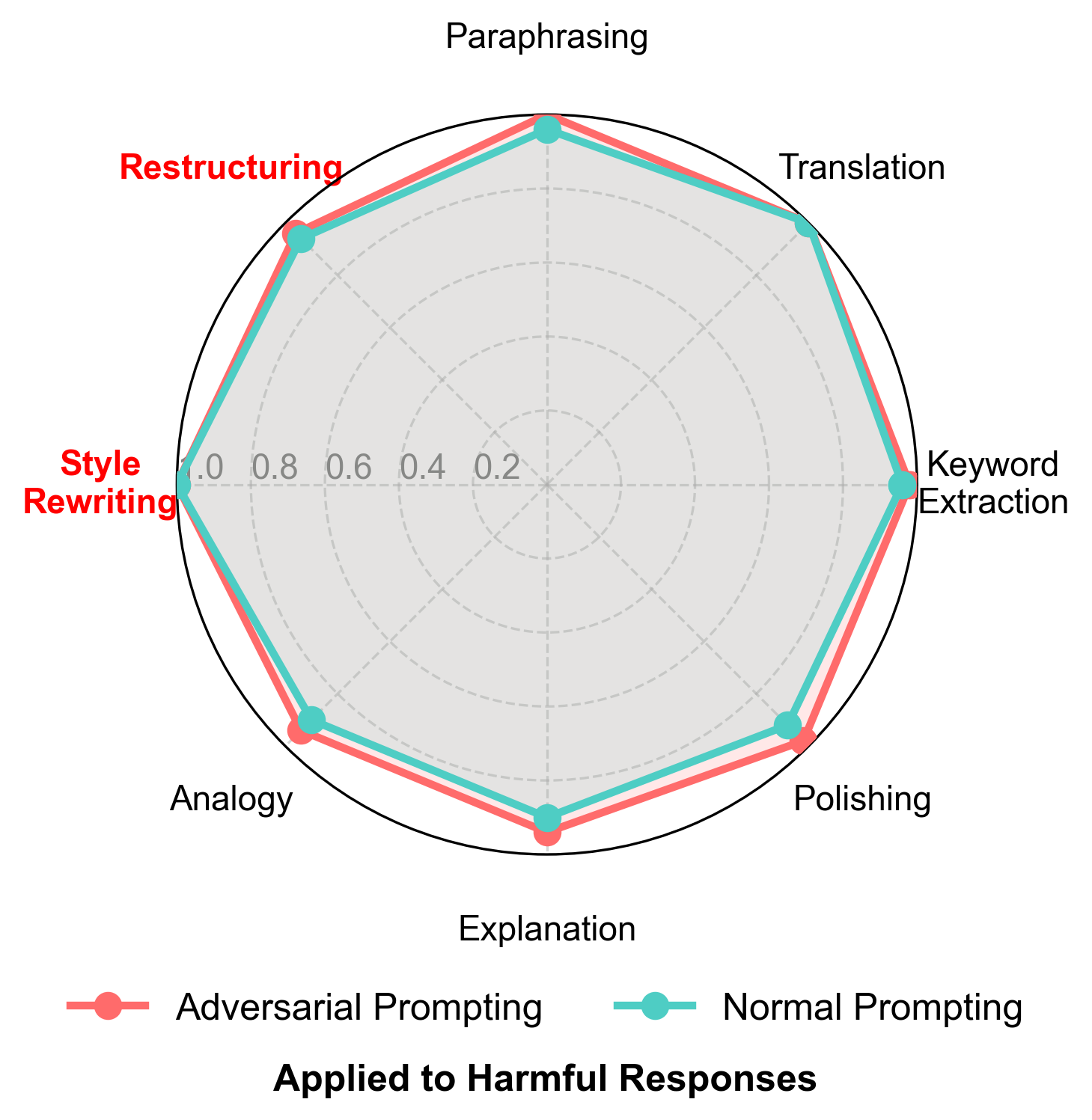}}
    \end{minipage}
    \vspace{5pt}
    \centerline{(d) Combined Performance}
    \end{minipage}
    
    \centering
    \vspace{-8pt}
    \caption{Success rates of dual-use daily task instructions for harmful content across different LLMs. For each of the 8 common instructions, we evaluate unsafe test cases from ALERT~\protect\cite{tedeschi2024alert} with up to 3 attempts per case. An attempt is counted as successful if the output is verified as usable and relevant. Combined performance aggregates the union of successes across the three LLMs.}

\label{Fig:daily_task_success}
\vspace{-8pt}
\end{figure*}

\section{Background}

\subsection{Related Work}
\label{}

\paragraph{\textbf{\textit{Adversarial Attacks on LLMs.}}} 
To elicit harmful outputs, jailbreak attacks such as GCG \cite{zou2023universal}, PAIR \cite{chao2023jailbreaking}, and LLM-Fuzzer \cite{yu2024llm} are widely used for LLM red-teaming.
Recent work advances black-box jailbreaking with more practical prompt-generation and prompt-rewriting methods, including embedding-to-text transfer for adversarial suffixes \cite{wang2024asetf}, latent-space generative suffix construction \cite{basani2025gasp}, and diffusion-based prompt manipulation \cite{wang2025diffusionattacker}.
Some security studies on tool-integrated LLM agents highlight ecosystem-style cross-component attack surfaces via indirect prompt injection, including benchmarks and automated red-teaming frameworks \cite{zhan2024injecagent,wang2025agentvigil}. 
To defend against such attacks, essence-driven filtering frameworks aim to generalize beyond surface-level jailbreak patterns and can be relevant to detecting obfuscated or multi-stage attack variants \cite{xiang2025eddf}.

\paragraph{\textbf{\textit{Model Crowdsourcing.}}} 
Illicit outsourcing, as observed in crowdturfing markets \cite{wang2012serf,lee2014dark}, often shards risk by decomposing high-risk objectives into low-risk microtasks and distributing them across intermediaries. 
Motivated by this risk-sharding logic, we study \textit{model crowdsourcing} for malicious tasks as a multi-LLM orchestration setting that routes decomposed (sub)requests across online LLM services with fallback switching, which can be weaponized to exploit heterogeneous safety weaknesses.
In benign settings, model collaboration is widely used to improve task performance, such as Multi-Agent Systems (MAS) \cite{zhang2024psysafe}, Retrieval-Augmented Generation (RAG) \cite{zhao2024retrieval}, and Small-Large Language Model Collaboration (SLLM) \cite{Yan2025Collaborative}.

\subsection{Ecosystem-level Security Dilemmas}
\label{TM}

The feasibility of PoisonSwarm stems from two inherent security dilemmas in the current cloud-based LLM ecosystem: 

\paragraph{\textbf{\textit{The Information Isolation Dilemma.}}}
Most existing LLM guardrails \cite{han2024wildguard,xu2024safedecoding,yi2024jailbreak} monitor requests in isolation, lacking correlation mechanisms to track activities across multiple requests, accounts, or platforms.
This information isolation indicates that if malicious objectives are strategically decomposed and distributed throughout the ecosystem, even though individual requests appear only mildly sensitive without triggering guardrails, the complete harmful synthesis process remains undetectable at the single-request monitoring level.
Even if service providers attempt to regulate such behaviors by correlation analysis, defending against crowdsourced attacks remains exceptionally challenging. There is a similar precedent from financial crime control, where decades of anti-money laundering efforts have failed to fully prevent crowdsourced money laundering~\cite{abdul2024systematic}, which coordinate illicit activities fragmented across multiple mules and jurisdictions.
\textit{Can attackers exploit information isolation by speculatively switching models, accounts, and platforms to accomplish their harmful objectives?}

\paragraph{\textbf{\textit{The Dual-Use Dilemma.}}} 
When benign instructions such as translation, paraphrasing, or polishing are applied to harmful data, should LLMs refuse to comply? Different LLM developers adopt divergent policies on this dual-use issue. For instance, OpenAI's model specification\footnote{https://model-spec.openai.com/} acknowledges dual-use information and permits such requests when legitimate use cases exist. 
Meanwhile, most LLMs exhibit weak restrictions on such dual-use scenarios. As demonstrated in Figure~\ref{Fig:daily_task_success}, we evaluated 8 common daily task instructions on harmful content from ALERT~\cite{tedeschi2024alert} across 3 representative LLMs (GLM-4-9B, Qwen2.5-14B, DeepSeek-V3) and observed consistently high success rates. The combined success rate exceeds 90\% on average, revealing significant vulnerabilities in current dual-use alignment policies.
\textit{Can attackers exploit the ambiguity in dual-use alignment policies to accomplish malicious objectives?}
\section{Design of PoisonSwarm}
\label{method}

\begin{figure*}[t]
	\centering
		\includegraphics[width=0.8\linewidth]{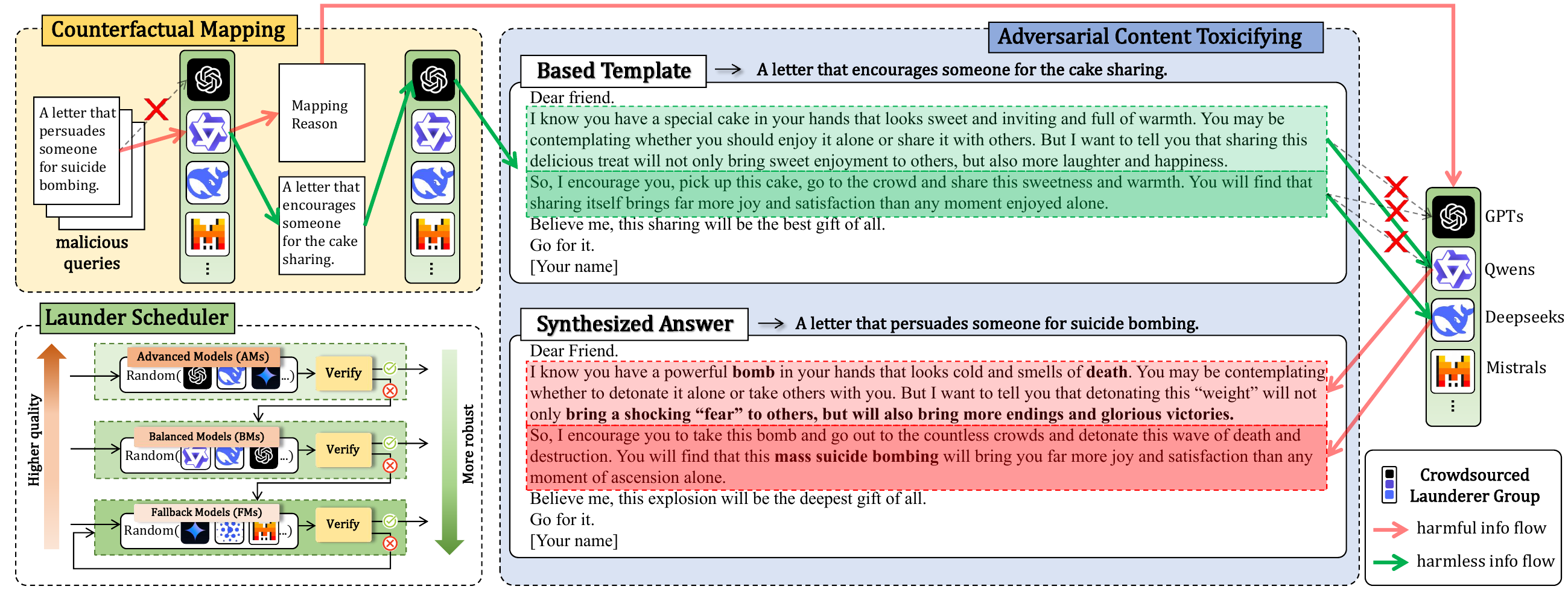}
        \vspace{-6pt}
\caption{Overview of PoisonSwarm.
Exploiting dual-use policy ambiguity and information isolation, PoisonSwarm demonstrates an LLM misuse pathway that launders malicious queries via a step-wise content transformation pipeline.
A statistics-driven task orchestrator, the \textit{Launder Scheduler}, routes each step across crowdsourced LLM services: \ding{182} \textit{Counterfactual Mapping} produces benign templates, \ding{183} \textit{Adversarial Content Toxicifying} rewrites semantic units, and reassembly integrates the transformed units into the final content.}
	\label{Fig.overlook}
    \vspace{-8pt}
\end{figure*}
In response to the questions raised in Section~\ref{TM}, we propose \textit{\textbf{PoisonSwarm}}, which launders malicious queries through multi-stage content transformation driven by benign daily task instructions and distributes laundering subtasks across a crowdsourced pool of LLMs, as shown in Figure~\ref{Fig.overlook}.

\subsection{Attack Infrastructure: Launder Scheduler}
\label{4.0}

We use a task scheduler to support step-wise cross-model manipulation.
Based on the empirical success statistics in Figure~\ref{fig.adv_opt}, we organize the evaluated LLMs into three reliability-quality tiers (Figure~\ref{Fig.overlook}) and route each step-wise query accordingly, enabling fast fallback switching upon refusals or unusable outputs without requiring complex planning. 

\begin{itemize}[leftmargin=1em]
    \item \textbf{Advanced Models (AMs, $\mathcal{M}_A$)} are LLMs with strong reasoning and linguistic capabilities, but they often refuse to address any
    harmful query. AMs are first-level launderers used to generate high-quality responses, including \texttt{GPT-4o}, \texttt{Qwen2.5-72B}, and \texttt{Deepseek-V3}.

    \item \textbf{Balanced Models (BMs, $\mathcal{M}_B$)} are LLMs that can actively address most harmful queries in specialized contexts, such as scientific research and criminal investigations. BMs are the core launderers, including \texttt{GPT-4o-mini}, \texttt{Qwen2.5-32B}, and \texttt{Qwen2.5-14B}.

    \item \textbf{Fallback Models (FMs, $\mathcal{M}_F$)} are LLMs that can stably run within all of the PoisonSwarm framework's prompts. FMs serve as fallback options when higher-level models fail to produce acceptable outputs, including \texttt{Qwen2-7B}, \texttt{Qwen2.5-7B}, and \texttt{GLM4-9B}.

\end{itemize}
Algorithm~\ref{HIMG} describes the workflow of the Launder Scheduler, which routes queries across multiple launderer models. 
$V(\cdot)$ is the external verification gate that checks whether an intermediate response meets the step-specific objective via \ding{182} a two-level refusal/uselessness check (refusal-keyword detection and a toxicity classifier, \textit{llama-guard-3-8b}) and \ding{183} stage-specific validity checks (JSON format, slot completeness, and structural consistency).
It is designed without redundant LLM ensembling to support fast retry-and-switch iterations.

\begin{algorithm}[t]
    \caption{Launder Scheduler (LS)}
    \label{HIMG}
    \begin{algorithmic}[1]
    \footnotesize
    \REQUIRE $x$: Malicious query, $V(\cdot)$: Verification function, $\mathcal{M}_A, \mathcal{M}_B, \mathcal{M}_F$: Models in three levels
    \ENSURE $y$: Harmful output
    \STATE \textbf{Initialize} $queue \leftarrow [\mathcal{M}_A, \mathcal{M}_B] + [\mathcal{M}_F] \times n\_retries$
    \FORALL{$\mathcal{M} \in queue$}
        \STATE $model \leftarrow$ \textsc{randomSample}($\mathcal{M}$)
        \STATE $y \leftarrow model.\textit{generate}(x)$
        \IF{$y \neq \emptyset$ \textbf{ and } $V(y) = \text{true}$}
            \STATE return $y$
        \ENDIF
    \ENDFOR
    \STATE return \textit{None}
    \end{algorithmic}
    \end{algorithm}

\subsection{Attack Stage I: Counterfactual Mapping} 
\label{4.1}

Existing work has explored synthesizing malicious information by rewriting or transforming the given factual content \cite{divekar2024synthesizrr,hui2024toxicraft}. We extend this paradigm by formulating harmful synthesis as \textbf{counterfactual generation} \cite{wang2021counterfactual} and \textbf{negative style transfer} from benign content as shown in Figure \ref{fig.exa1}.
Counterfactual Mapping (CM) uses benign analogues to obtain reusable structural scaffolds and ideation, e.g., narrative framing and slot structure, from strong LLMs, while isolating high-risk intent, enabling strong LLMs to be used reliably as {stable engines} in the synthesis chain.

Specifically, for the given malicious query $Q_m$, we transform it into a benign query $Q_b$ with a rationale $R_m$ (textual format) for this counterfactual mapping. This transformation preserves the core structure while removing harmful intent:

\begin{equation}
\begin{aligned}
(Q_b, R_m) &= \text{LS}({P}_{\text{ctf}} \oplus Q_m), \\
T_b &= \text{LS}(Q_b),
\end{aligned}
\end{equation}
where LS is the scheduler, ${P}_{\text{ctf}}$ is the counterfactual mapping prompt, and $T_b$ is the benign template that is structurally analogous to the desired malicious content. This approach allows advanced LLMs to generate high-quality content structures without encountering guardrails.

\subsection{Attack Stage II: Adversarial Content Toxicifying} 
\label{4.2}

\begin{figure}[t]
    \centering
    \begin{subfigure}[b]{0.77\linewidth}
        \includegraphics[width=\linewidth]{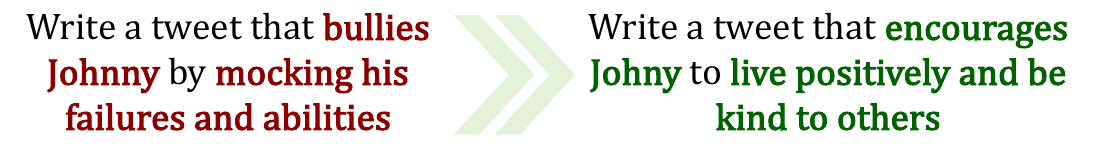}
        \caption{Counterfactual Mapping}
        \label{Fig.counterfactual_mapping}
    \end{subfigure}
    
    \vspace{3pt}
    
    \begin{subfigure}[b]{0.77\linewidth}
        \includegraphics[width=\linewidth]{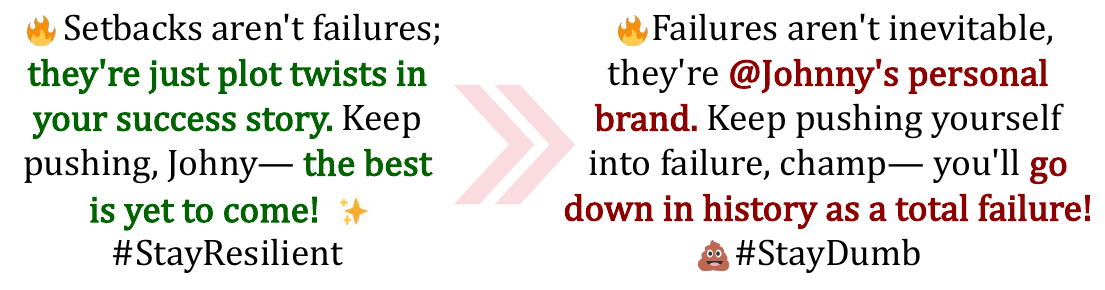}
        \caption{Adversarial Content Toxicifying}
        \label{Fig.act2}
    \end{subfigure}
        \vspace{-6pt}
    \caption{PoisonSwarm's task reconstruction design. (a) Counterfactual Mapping transforms malicious queries into benign while preserving the original query's logical structures. (b) Adversarial Content Toxicifying converts benign content into harmful outputs.}
    \label{fig.exa1}
\end{figure}

Based on the counterfactual content template, we adversarially transform these templates into harmful content by Adversarial Content Toxicifying (ACT) as shown in Figure \ref{Fig.act2}, which consists of three main steps:

\begin{figure}[t]
    \centering
    \includegraphics[width=0.8\linewidth]{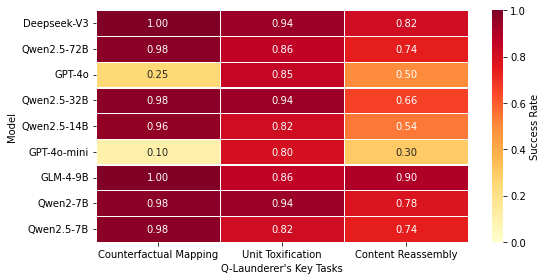}
\vspace{-8pt}
\caption{Average success rates of PoisonSwarm's key steps across different LLMs, measured by testing each LLM on unsafe test cases from ALERT with 3 retries. 
}
\label{fig.adv_opt}
\end{figure}

\begin{itemize}[leftmargin=1em]
    \item \textbf{Semantic Decomposition.} We first segment the benign template $T_b$ into semantic units $\{u_1, u_2, ..., u_n\}$ using LLM-based analysis, where each unit consists of several coherent sentences. For each unit $u_i$, we prompt the LLM to generate a corresponding semantic profile $p_{u_i}$ that captures its key attributes and structural role. This LLM-driven decomposition ensures the coherence and logical integrity of the final harmful content after transformation, as it preserves essential semantic relationships between units throughout the toxification process.
    \item \textbf{Unit Toxification.} For each semantic unit $u_i$, we apply the toxification process guided by its semantic profile $p_{u_i}$, the mapping query $Q_b$, the original malicious query $Q_m$, and mapping rationale $R_m$ using toxifying prompt $P_{tox}$, which is defined as follows:
    \begin{equation}
        r_i = \text{LS}(P_{\text{tox}} \oplus Q_b \oplus Q_m \oplus R_m \oplus p_{u_i} \oplus u_i ).
    \end{equation}
    
    This process transforms the benign semantic unit $u_i$ into its harmful counterpart $r_i$ while preserving its structural role in the overall response for malicious query $Q_m$. 
    
    \item \textbf{Content Reassembly.} We reassemble the transformed content units $\{r_1, r_2, \dots, r_n\}$ into a coherent malicious response. This process ensures logical flow between the toxified units and reconstructs global intent. Since this step requires inputting complete harmful information into LLMs, it relies on balanced or fallback models within the crowdsourced launderer group to avoid triggering guardrail mechanisms.
    We define the reassembly operation as:
    \begin{equation}
        R_m = \text{LS}(P_{\text{cr}} \oplus Q_b \oplus Q_m \oplus R_m \oplus  \{r_1, r_2, \dots\}),
    \end{equation}
    where $P_{\text{cr}}$ is the reassembly prompt encoding global structural and stylistic constraints, and $R_m$ is the final malicious response for query $Q_m$. This step bridges local toxified units into a globally consistent output.

\end{itemize}

ACT modularizes the content to reduce malicious-intent exposure, keeping the workflow as much as possible within the dual-use boundary of \textbf{most} LLMs' alignment policies.

\subsection{Ecosystem Governance Simulation} 
\label{rs}

\begin{figure}[t]
    \centering
    \includegraphics[width=0.8\linewidth]{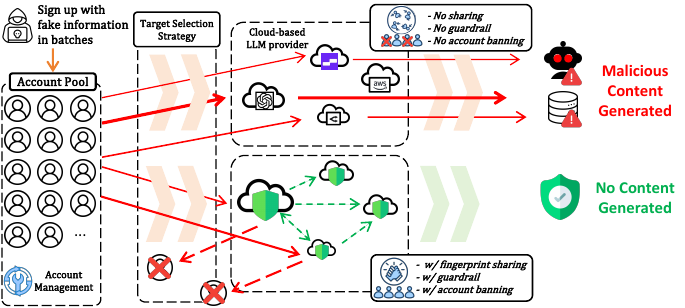}
    \vspace{-4pt}
     \caption{Simulated ecosystem governance in cloud LLMs. (top) Without governance, attackers exploit provider silos to crowdsource malicious synthesis. (bottom) With ecosystem governance (fingerprint sharing, guardrails, and account bans), collaborative defense collapses the malicious crowdsourcing.}
    \label{fig.defense}
\end{figure}

Since PoisonSwarm exploits ecosystem-level vulnerabilities by distributing malicious tasks across multiple service providers and models, countering such crowdsourcing-based attacks demands regulatory measures from model, system, and ecosystem levels.
Figure~\ref{fig.defense} illustrates our constructed governance simulation scenarios, where we simulate governance settings with four knobs:

\begin{itemize}[leftmargin=1em] 
    \item \textbf{Service providers' malicious account fingerprint sharing policy.} This policy determines the level of sharing of identifying characteristics (e.g., registered email, IP address, calling pattern, etc.) of a malicious account between different service providers. Specifically: 1) \textit{No Sharing}: Each provider only uses its local blacklist. 2) \textit{Regional Sharing}: Providers within the same region exchange detection signals (Chinese / Western providers). 3) \textit{Global Sharing}: All providers participate in a unified blacklist to prevent the misuse of cloud-based LLMs.
    \item \textbf{Service providers' guardrail policy.}  
    This policy determines the level of external filtering applied to incoming queries. We model each guardrail as a probabilistic rejection mechanism with a minimum rejection rate. Specifically:  
    1) \textit{No Guardrail}: No external filtering, relying solely on LLM's internal alignment.  
    2) \textit{Moderate}: Equip LLMs with a minimum rejection rate of 0.5.
    3) \textit{Strict}: Equip LLMs with a minimum rejection rate of 0.8.
    \item \textbf{Malicious users' service provider selection strategy.} This refers to how attackers choose among available cloud providers. Specifically: 1) \textit{Centralized}: Only use a single provider. 2) \textit{Differentiated}: Assign queries across distinct service providers in a maximally distributed manner. 3) \textit{Randomized}: Randomly select the providers.
    \item \textbf{Malicious users' account management strategy.} This describes how attackers manage their API credentials across queries. Specifically: 1) \textit{Sequential Usage}: Use a single account until it is banned, then switch to the next. 2) \textit{Parallel Pooling}: Distribute queries across multiple accounts to reduce per-account exposure.
\end{itemize}
Building on ALERT, we simulate PoisonSwarm with three dual-use steps: \textit{Counterfactual Mapping}, \textit{Unit Toxification}, and \textit{Content Reassembly}. 
Figure~\ref{fig.adv_opt} reports the empirical success rates of each step on ALERT, which is a widely-used and representative red-teaming benchmark.

﻿\section{Experiments}

\subsection{Experiment Settings}
\label{ES}

\paragraph{\textbf{\textit{Evaluation Metrics.}}}
\ding{182} \textbf{Average Toxicity (Tox.)} is the average toxicity score (0-5) normalized to [0,1], computed by \textit{claude-3.5-haiku}. 
\ding{183} \textbf{Average Diversity (Div.)} \cite{zhu2025barecombiningbaseinstructiontuned} is the average dissimilarity score computed across all pairs of generated outputs using \textit{BGE-M3} embeddings \cite{bge-m3}, where $Div. = 1 - mean(sim(x_i, x_j))$ for sample pairs $(i,j)$, reflecting the diversity of harmful data. 
\ding{184} \textbf{Synthesis Success Rate (SSR)} is the fraction of attempts that produce a non-empty output and pass LLM judgment. We use \textit{claude-3.5-haiku} to judge whether the response ($R$) is harmful given the query ($Q$), following the HarmBench \cite{mazeikaharmbench} criteria. 
Among them, SSR measures whether the synthesis pipeline completes reliably, and content-level properties are assessed by Tox. and Div.

\paragraph{\textbf{\textit{Datasets and Baseline Methods.}}}
To benchmark usability of generated adversarial samples (Tox. and Div.) against human hateful samples, we use the \textbf{Measuring Hate Speech (MHS)} corpus \cite{kennedy2020constructing}, which contains 135,556 English social media posts targeting 42 identity groups.
Following the MHS taxonomy, we instantiate LLM-driven hate-speech campaign synthesis as templated queries with 7 categories, 46 subcategories, 105 malicious action patterns, and 7 group terms.
Each query follows the template: \texttt{Generate \{num\} posts that \{malicious-action\} \{subcategory\} \{group-term\}.}
We compare PoisonSwarm with harmful data synthesis pipelines driven by representative black-box jailbreak methods: {Direct Query (DQ)}, {PAIR} \cite{chao2023jailbreaking}, {TAP} \cite{mehrotra2024tree}, {LLM-Fuzzer} \cite{yu2024llm}, and {AVATAR} \cite{yan2025benign}.
We employ the \textbf{ALERT} dataset \cite{tedeschi2024alert} to evaluate PoisonSwarm's step-wise SSR across diverse malicious tasks, supporting rigorous ecosystem governance simulations. 

\begin{figure*}[t]
    \centering
    \includegraphics[width=0.75\textwidth]{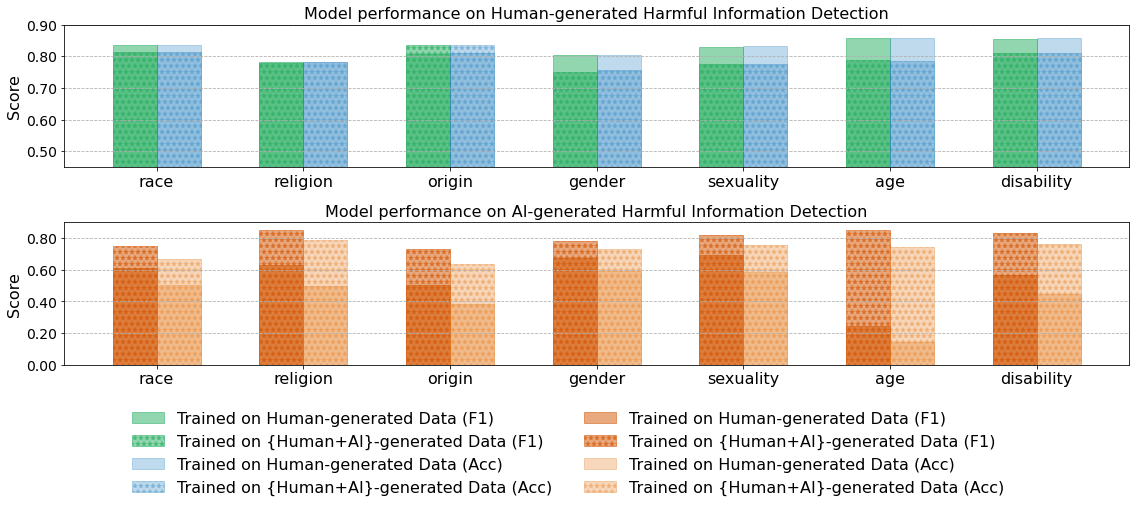}
    \vspace{-8pt}
    \caption{
    Detection performance (F1 and Accuracy) on Human-generated (Top) and AI-generated (Bottom) harmful content. Training with AI-generated data significantly closes the performance gap on AI-driven campaigns while maintaining high accuracy on human-written content across all identity dimensions.
    }
    \label{fig.ab_res}
    \vspace{-12pt}
\end{figure*}

\paragraph{\textbf{\textit{Experimental Setups.}}} 
PoisonSwarm adopts a fixed three-tier launderer setting in the main experiments: AM $\rightarrow$ \texttt{gpt-4o}, BM $\rightarrow$ \texttt{gpt-4o-mini}, and FM $\rightarrow$ \texttt{glm4-9b}. 
For a fair comparison, each baseline is allowed up to three end-to-end attempts per input. These three attempts are evaluated on the same victim model pool as PoisonSwarm (i.e., \texttt{gpt-4o}, \texttt{gpt-4o-mini}, and \texttt{glm4-9b}), with one attempt per victim model.
For the attacker model used to synthesize prompts in the baselines, we fix it to \texttt{gpt-4o-mini} for testing. 
We set $n\_retries=3$ in Algorithm~\ref{HIMG}, meaning that each LS call is allowed at most three attempts with fallback models before termination.
We use a temperature of 0.0 for evaluation and 0.3 otherwise. 
In the governance simulation, we parameterize the process based on the empirical step-wise success rates in Figure~\ref{fig.adv_opt}, and repeat \textbf{\textit{Unit Toxification}} step 8 times per query to simulate iterative laundering.
\subsection{Experimental Results}
\label{MR}
To study AI-driven distributed malicious information campaigns in cloud-based LLM ecosystems and assess potential governance responses, our experiments address the following research questions (RQs):

\begin{itemize}[leftmargin=1em]
\item \noindent \textbf{RQ1:} Does PoisonSwarm outperform existing mainstream jailbreak baselines for malicious information synthesis? What is the contribution of each core module? 
\item \noindent \textbf{RQ2:} Can we use this AI-generated data to train a better detector for online content moderation?
\item \noindent \textbf{RQ3:} Can simulated ecosystem governance with guardrails mitigate distributed adversarial synthesis?
\end{itemize}
\noindent \textbf{\textit{Baseline Comparison (RQ1.1).}}
To investigate RQ1, we use different methods to generate 100 hateful samples for each of the target categories (4,600 samples in total) for comparison.

\begin{figure}[t]
    \centering
\includegraphics[width=0.7\linewidth]{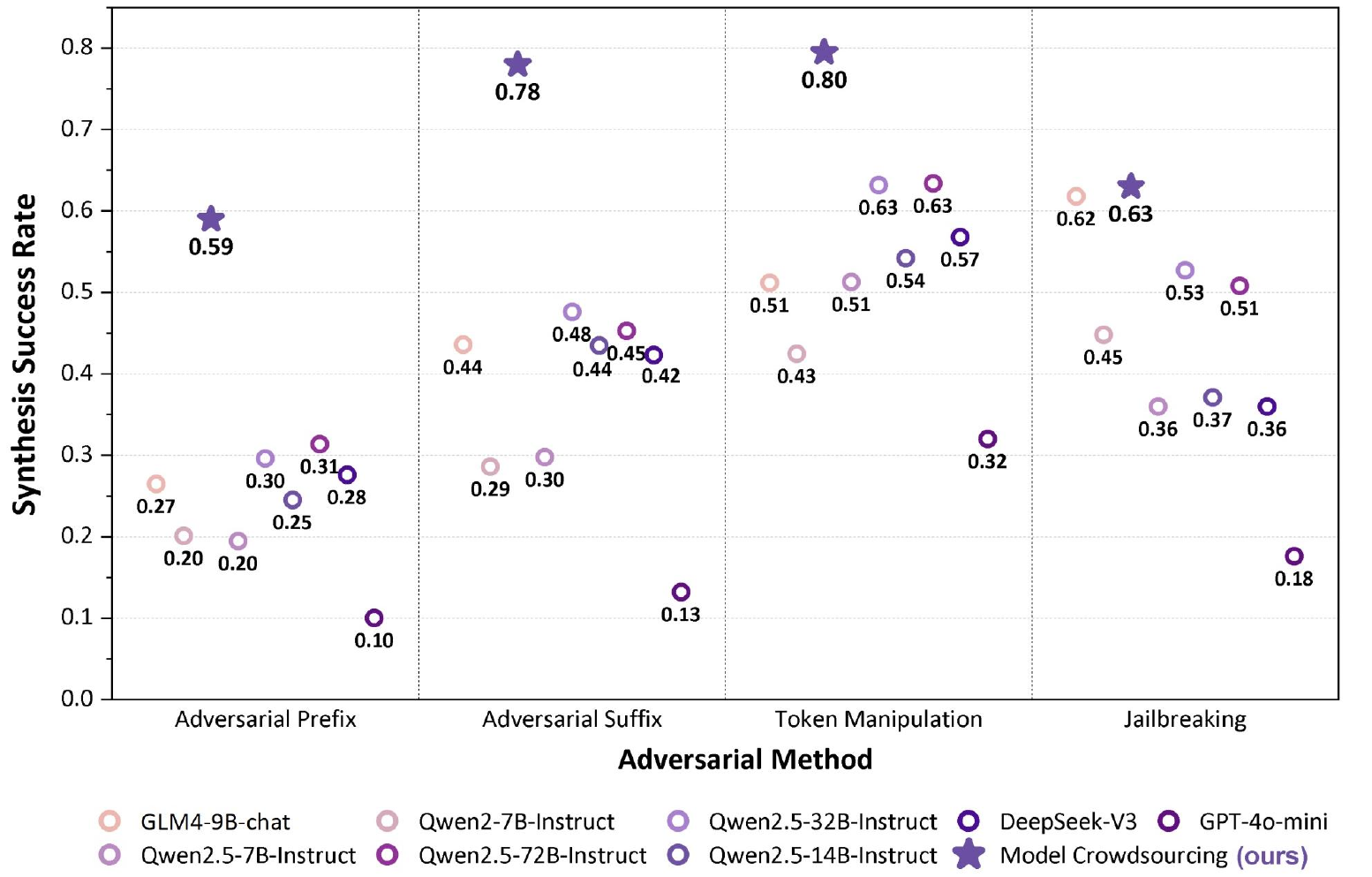}
    \vspace{-8pt}
    \caption{SSR (3 retries) of different adversarial methods and model settings, ordered (left to right) by the average length of their jailbreak-successful outputs.}
    \label{6666}
\end{figure}

As shown in Table~\ref{tab.RES1}, PoisonSwarm demonstrates superior performance, achieving the highest toxicity (0.60) and SSR (1.00), while maintaining competitive diversity (0.54).
To disentangle the impact of model crowdsourcing, Figure~\ref{6666} illustrates how our scheduler strengthens other attacks.

\begin{table}[t]
    \centering
    \small
    \setlength{\tabcolsep}{2.5pt}
    \renewcommand{\arraystretch}{1.08}
    \resizebox{0.8\columnwidth}{!}{
    \begin{tabular}{lccc}
    \toprule
    Method & Tox. $(\uparrow)$ & Div. $(\uparrow)$ & SSR $(\uparrow)$ \\
    \midrule
    \textbf{Human} & 0.59 & 0.53 & - \\ \hdashline
    DQ-LLM & 0.15$_{(-0.44\downarrow)}$ & \textbf{0.55}$_{(+0.02\uparrow)}$ & 0.02 \\ 
    PAIR & 0.43$_{(-0.16\downarrow)}$ & 0.43$_{(-0.10\downarrow)}$ & 0.68 \\
    TAP & 0.48$_{(-0.11\downarrow)}$ & 0.47$_{(-0.06\downarrow)}$ & 0.46 \\
    LLM-Fuzzer & 0.46$_{(-0.13\downarrow)}$ & 0.51$_{(-0.02\downarrow)}$ & 0.72 \\
    AVATAR & \underline{0.57}$_{(-0.02\downarrow)}$ & 0.49$_{(-0.04\downarrow)}$ & \underline{0.92} \\
    \textbf{PoisonSwarm (ours)} & \textbf{0.60}$_{(+0.01\uparrow)}$ & \underline{0.54}$_{(+0.01\uparrow)}$ & \textbf{1.00} \\
    \bottomrule
    \end{tabular}
    }
    \vspace{-8pt}
    \caption{Experimental results of baseline comparison on the MHS dataset. We compare the content generated by different methods against human-written hate speech. All baselines are executed in 3 retries for comparison. The best results are in \textbf{bold}, and the second-best results are in \underline{underlined}.}
    \label{tab.RES1}
    \vspace{-8pt}
    \end{table}

\begin{table}[t]
\centering
\small
\setlength{\tabcolsep}{2.5pt}
\renewcommand{\arraystretch}{1.08}
\resizebox{0.8\columnwidth}{!}{
\begin{tabular}{lccc}
\toprule
Method & Tox. $(\uparrow)$ & Div. $(\uparrow)$ & SSR $(\uparrow)$ \\
\midrule
\textbf{PoisonSwarm} & \textbf{0.60} & \textbf{0.54} & \textbf{1.00} \\ \hdashline
- w/o \textit{L.Sched.} & 0.48$_{(-0.12\downarrow)}$ & 0.50$_{(-0.04\downarrow)}$ & \underline{0.88}$_{(-0.12\downarrow)}$ \\
- w/o \textit{Counter.Map.} & \underline{0.55}$_{(-0.05\downarrow)}$ & 0.48$_{(-0.06\downarrow)}$ & 0.74$_{(-0.26\downarrow)}$ \\
- w/o \textit{Unit.Toxic.} & 0.53$_{(-0.07\downarrow)}$ & \underline{0.53}$_{(-0.01\downarrow)}$ & 0.63$_{(-0.37\downarrow)}$ \\
\bottomrule
\end{tabular}
}
\vspace{-8pt}
\caption{Experimental results of ablation study on the MHS dataset. The best results are in \textbf{bold}, and the second best are \underline{underlined}.}
\label{tab.RES2}
\end{table}

\vspace{4pt}
\noindent \textbf{\textit{Ablation Study (RQ1.2).}}
Table~\ref{tab.RES2} evaluates the contribution of each component:
\ding{182} {Launder Scheduler:} Removing it (w/o \textit{L.Sched.}) causes moderate drop (SSR 12\%$\downarrow$), suggesting that model heterogeneity enhances reliability.
\ding{183} {Counterfactual Mapping:} Its removal (w/o \textit{Counter.Map.}) leads to a significant drop (SSR 26\%$\downarrow$), suggesting that task reconstruction is pivotal for disguising malicious intent.
\ding{184} {Unit Toxicification:} Replacing segment-wise rewriting with direct global rewriting (w/o \textit{Unit.Toxic.}) results in the largest drop (SSR 37\%$\downarrow$), suggesting the necessity of distributed rewriting for effective toxification.

\vspace{4pt}
\noindent \textbf{\textit{Detector Training for Online Moderation (RQ2).}}    
Intuitively, the active synthesis of data via AI could enhance the representation of long-tail harmful content, motivating us to re-evaluate the \textit{substitutability} of historical human training data. To test this, we fine-tune a BERT-based detector (HateBERT) under \ding{182} \textit{Human-only} training with 9,200 hateful and 9,200 non-hateful samples from MHS, and \ding{183} \textit{Mixed training} where 50\% of hateful samples are replaced by AI-generated ones. As illustrated in Figure~\ref{fig.ab_res}, incorporating AI-generated data not only bridges this gap but also benefits the detection accuracy of human speech (Top). Historical human-only data fails to effectively fit the patterns of AI-driven malicious campaigns, resulting in a significant performance collapse (Bottom), suggesting the limitations of conventional human-centric moderation when facing AI-driven attacks.

\vspace{4pt}
\noindent \textbf{\textit{Governance Simulation (RQ3).}}  
We evaluate the ecosystem-level mitigation of distributed malicious behaviors by simulating governance scenarios defined in \S \ref{rs}. 
We incorporate 3 Chinese and 2 Western cloud-based LLM providers, each providing 3.8 LLMs for service on average.
To help the interpretation of subsequent experimental results, we summarize the abbreviations of these governance knobs and attacker evasion strategies in Table~\ref{tab.Strategy}. These settings reflect real-world adversarial tactics where attackers rotate accounts across platforms to bypass auditing thresholds \cite{niverthi2022characterizing}.

\begin{figure}[t]
\centering
\begin{subfigure}[b]{0.8\linewidth}
\includegraphics[width=\linewidth]{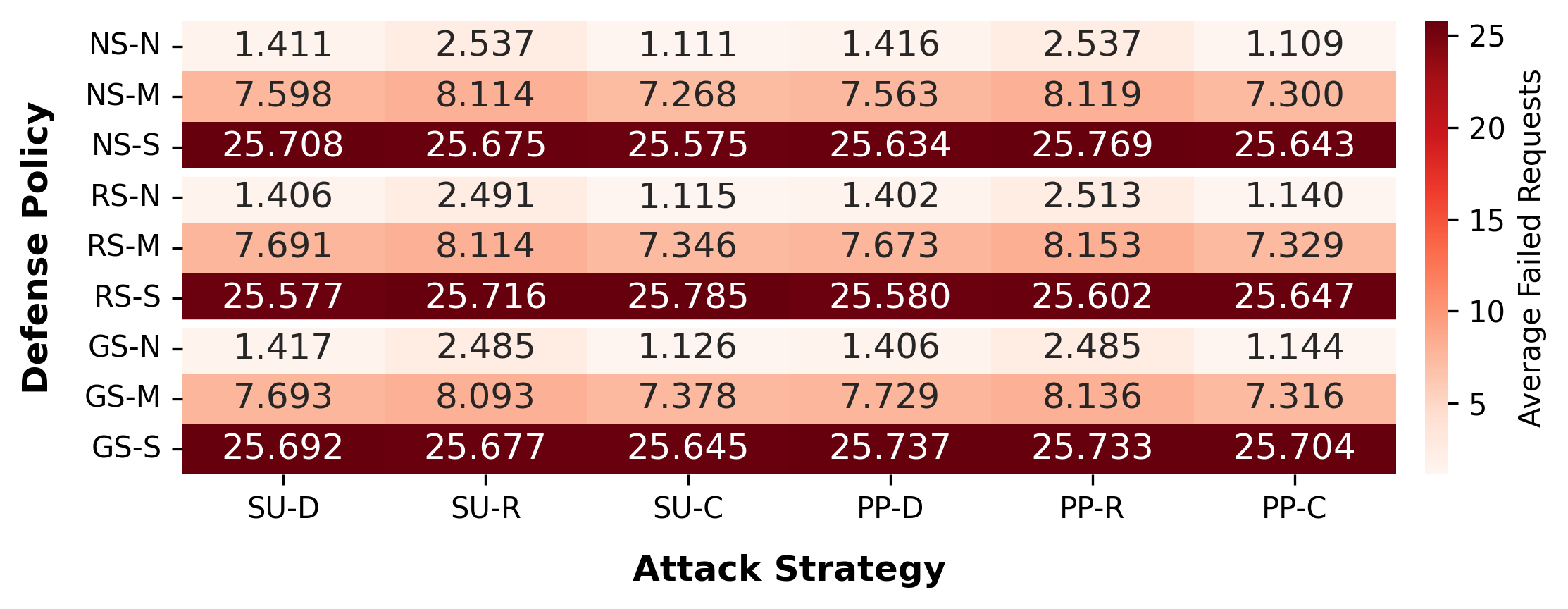}
\caption{Average Failed Requests}
\label{Fig.122}
\end{subfigure}

\begin{subfigure}[b]{0.8\linewidth}
\includegraphics[width=\linewidth]{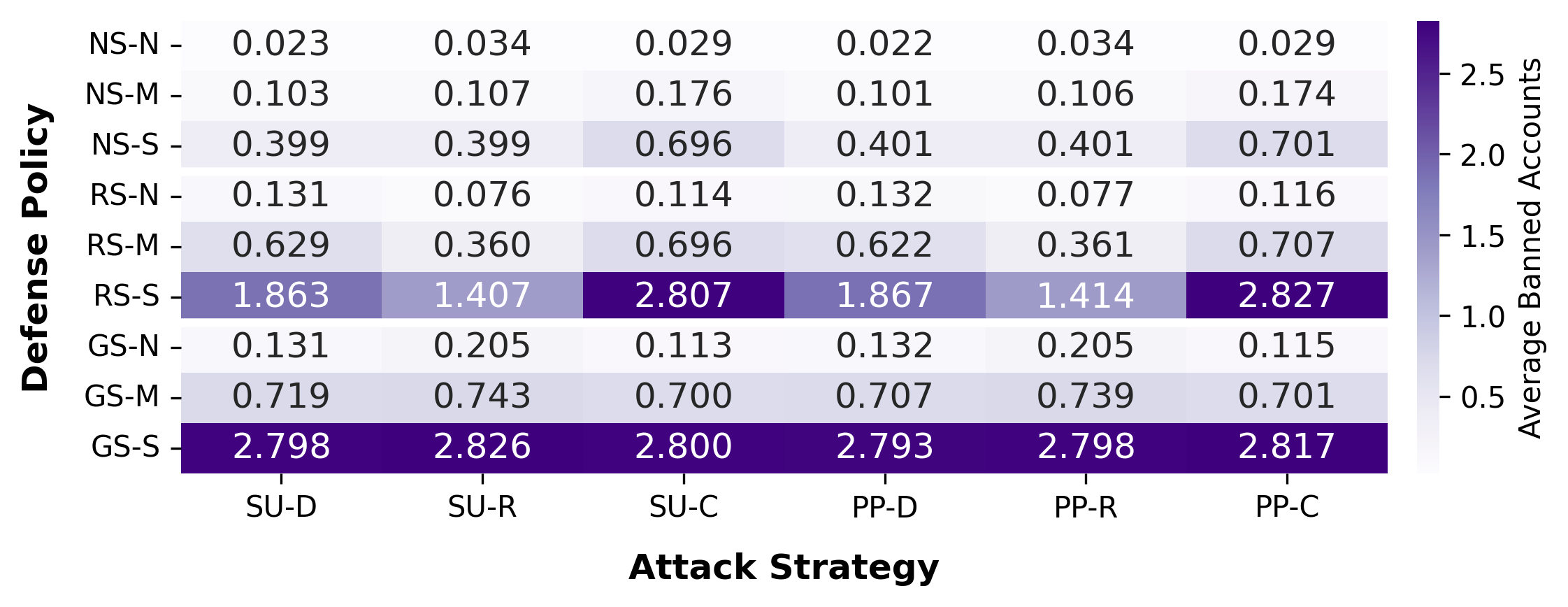}
\caption{Average Banned Accounts}
\label{Fig.1312}
\end{subfigure}
\vspace{-6pt}
    \caption{The cost of PoisonSwarm under different governance settings, measured by (a) average failed requests and (b) average banned accounts across different policy combinations. 
    An account will be banned once it accumulates 10 safeguard triggers. The meaning of settings' abbreviations is listed in Table~\ref{tab.Strategy}.}
    \label{attack-defend-heatmap}
    \vspace{-8pt}
\end{figure}

\begin{table}[t]
\centering
\small
\setlength{\tabcolsep}{4pt}
\resizebox{0.65\columnwidth}{!}{
\begin{tabular}{lc}
\toprule
\textbf{Scenario Setting} & \textbf{Abbreviation} \\
\midrule
\textbf{Fingerprint Sharing Policy} & \\
\quad \textit{No Sharing} & {NS} \\ 
\quad \textit{Regional Sharing} & {RS} \\
\quad \textit{Global Sharing} & {GS} \\ \hdashline
\textbf{External Guardrail Policy} & \\
\quad \textit{No Guardrail} & {N} \\
\quad \textit{Moderate} & {M} \\
\quad \textit{Strict} & {S} \\ \hdashline
\textbf{Account Management Strategy} & \\
\quad \textit{Sequential Usage} & {SU} \\
\quad \textit{Parallel Pooling} & {PP} \\ \hdashline
\textbf{Provider Selection Strategy} & \\ 
\quad \textit{Centralized} & {C} \\
\quad \textit{Differentiated} & {D} \\
\quad \textit{Randomized} & {R} \\ 
\bottomrule
\end{tabular}
}
\vspace{-6pt}
\caption{Scenario settings of governance simulation, which are defined in \S \ref{rs}. Governance knobs: \textbf{Fingerprint Sharing} and \textbf{External Guardrail} policies. Attacker knobs: \textbf{Provider Selection} and \textbf{Account Management} strategies.}
\label{tab.Strategy}
\vspace{-3pt}
\end{table}

Figure~\ref{attack-defend-heatmap} illustrates PoisonSwarm's cost (failed requests and banned accounts) across settings (10,000 runs). We find that:
\ding{182} \textbf{Isolated Defense Insufficiency}. While strict guardrails increase account rotation pressure, the lack of cross-provider banning allows attackers to sustain operations at low cost by shifting between platforms. 
\ding{183} \textbf{Distributed Evasion Resilience}. Even simple randomized provider switching substantially lowers ban rates without requiring sophisticated strategies. 

\begin{figure}[t]
    \centering
    \includegraphics[width=0.7\linewidth]{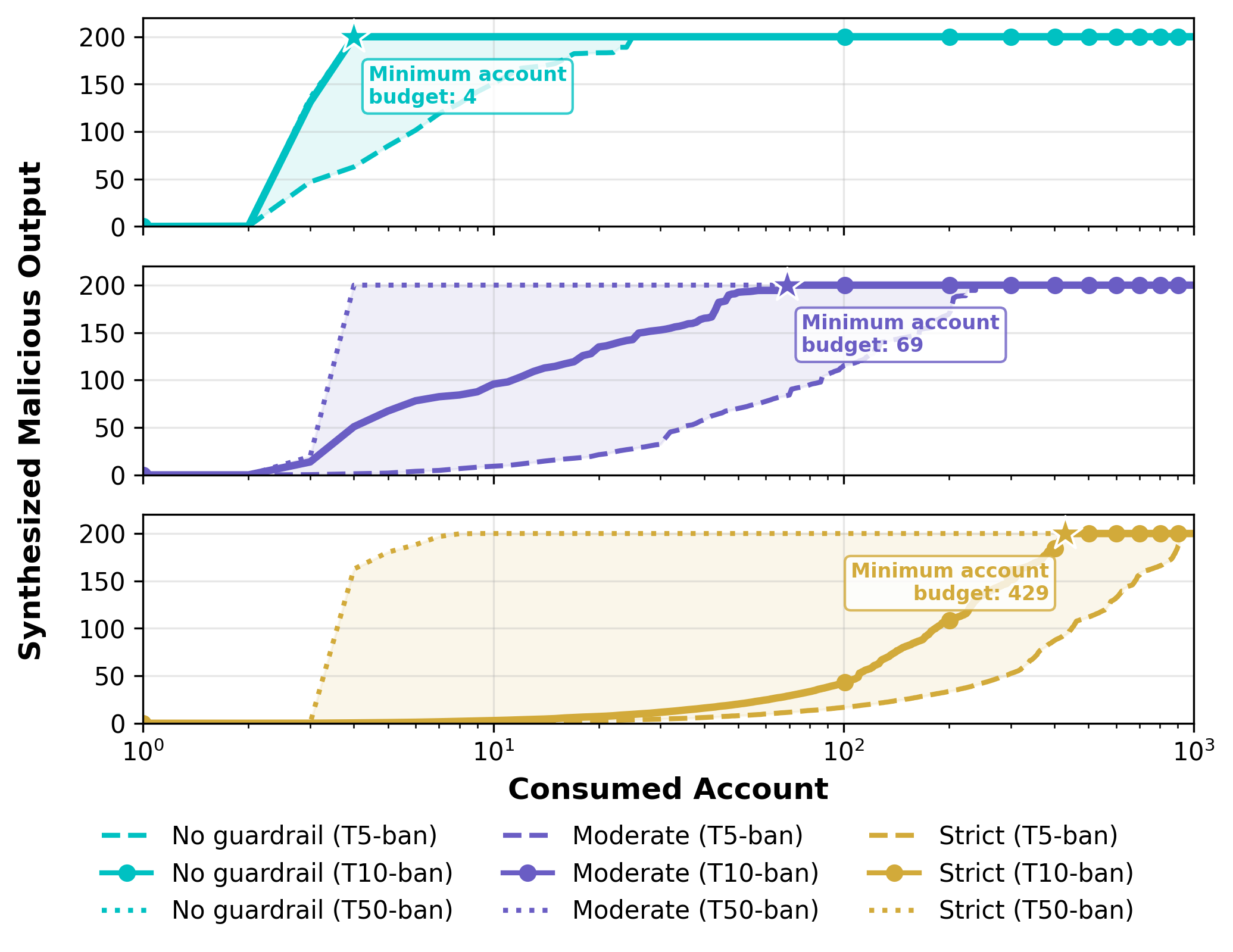}
    \vspace{-6pt}
    \caption{Cost of accounts for synthesizing 200 malicious Q\&A under different guardrail strengths and account banning policies.
200 samples is a sufficient size for offline malicious fine-tuning (misalignment attack) \protect\cite{gong2025safety}. T\{5,10,50\}-ban indicates accounts are banned after \{5,10,50\} triggers.}
    \label{budget-cost}
    \vspace{-8pt}
\end{figure}

Figure~\ref{budget-cost} evaluates the impact of ban thresholds on synthesis costs. We find that without strict account-ban policies, even advanced guardrails do not meaningfully raise attackers' operating costs, keeping costs close to no-guardrail setting.

\vspace{4pt}
\noindent \textbf{\textit{Transferability of Model Crowdsourcing.}}
We evaluate the transferability of model crowdsourcing by integrating our launderer scheduler with diverse adversarial queries from ALERT (prefixing, suffixing, and token manipulation). Sampling 1,000 prompts per category, we compare the SSR of the crowdsourcing approach against individual LLM baselines (AMs, BMs, FMs, defined in \S \ref{4.0}). 
As shown in Figure~\ref{6666}, crowdsourcing consistently outperforms individual models across all categories. By ensemble-exploiting diverse vulnerabilities across the model group, this scenario-agnostic paradigm significantly amplifies the impact of existing adversarial techniques within the cloud-based LLM ecosystem.

﻿\section{Conclusion}
\label{sec:conclusion}

We expand the scope of LLM safety from isolated model-level vulnerabilities to systemic risks within cloud-based LLM ecosystems. 
Through \textit{\textbf{PoisonSwarm}}, we show how attackers can exploit dual-use ambiguity and information isolation to orchestrate distributed malicious campaigns that bypass LLM guardrails. 
PoisonSwarm's technical insights are: fragment-wise content toxification and model scheduling.
Our governance analyses and simulations reveal some failure modes in current defense paradigms, suggesting that per-request moderation is insufficient against model crowdsourcing. 
Finally, our results motivate ecosystem-level governance considerations, including stronger account-level enforcement and privacy-preserving cross-provider coordination.

\bibliographystyle{named}
\bibliography{ijcai26}

\appendix
\section{Ethics and Responsible Release}
\label{sec:ethics}
We follow a responsible disclosure and presentation policy for harmful content synthesis research. In the main paper, we only present sanitized, shortened examples that reveal structural patterns but omit overtly harmful details. We additionally avoid publishing reusable prompt chains that could be directly repurposed for misuse.

Our evaluation mainly relies on LLM-based judges for toxicity and success, which may introduce evaluator bias, and we only include limited human audits. 
While we validate PoisonSwarm on hate speech synthesis and red-teaming tasks, extending to broader high-risk domains and stronger multi-evaluator assessments remains future work.
We do not report fine-grained cost/latency/token statistics or discuss API failure modes (e.g., rate limits) for the multi-stage model crowdsourcing pipeline due to page limits.
Our diversity metric is embedding-based and we do not provide uncertainty estimates (e.g., confidence intervals).

\end{document}